\pdfoutput=1
\documentclass[conference]{IEEEtran}
\IEEEoverridecommandlockouts

\usepackage{amsmath}
\usepackage{amssymb}
\usepackage{mathtools}
\usepackage{graphicx}
\usepackage{booktabs}
\usepackage{multirow}
\usepackage[table]{xcolor}
\usepackage{subcaption}
\usepackage{xspace}
\usepackage{microtype}
\usepackage{url}
\usepackage{algorithm}
\usepackage[noend]{algpseudocode}

\usepackage{hyperref}
\hypersetup{
  colorlinks=true,
  linkcolor=blue!60!black,
  citecolor=blue!60!black,
  urlcolor=blue!60!black,
  breaklinks=true,
  pdftitle={Discovery of Diverse and Realistic Financial Tail-Risk Using Generative Flow Networks},
  pdfauthor={Adir Serruya and Gilad Katz},
  pdfkeywords={generative flow networks, financial risk, tail-risk discovery, scenario generation, stress testing}
}
\usepackage[capitalize,noabbrev]{cleveref}

\newcommand{\MethodName}{GRID\xspace}

\definecolor{highlight21}{HTML}{dde5b6} 
\definecolor{highlight08}{HTML}{FFE4B2} 
\definecolor{highlight02}{HTML}{C9D7F8} 

\begin{document}

\title{Discovery of Diverse and Realistic Financial Tail-Risk Using Generative Flow Networks}

\author{
\IEEEauthorblockN{Adir Serruya}
\IEEEauthorblockA{\textit{Faculty of Computer and Information Science}\\
\textit{Ben-Gurion University of the Negev}\\
Beer-Sheva, Israel\\
adirser@post.bgu.ac.il}
\and
\IEEEauthorblockN{Gilad Katz}
\IEEEauthorblockA{\textit{Faculty of Computer and Information Science}\\
\textit{Ben-Gurion University of the Negev}\\
Beer-Sheva, Israel\\
giladkz@bgu.ac.il}
}

\maketitle

\begin{abstract}
Accurate modeling of rare, high-impact financial events is essential for robust risk management. Sampling-based techniques such as Sequential Markov Chain Monte Carlo (SMCMC) generate diverse scenarios but inefficiently explore high-risk regions, while Deep Reinforcement Learning (DRL) converges on narrow sets of scenarios, limiting diversity. We present \MethodName, which uses Generative Flow Networks (GFlowNets) -- samplers that draw objects in proportion to a reward -- to systematically generate diverse and realistic financial tail-risk scenarios. Building on continuous GFlowNets, \MethodName introduces a flexible mixture-policy architecture with variable-specific distribution families for high-dimensional, sequential macro-financial data, together with plausibility constraints that keep generated trajectories realistic. A central practical advantage is robustness to reward-proxy imperfection: because GFlowNets sample proportionally rather than maximize, they remain broadly exploratory even when the oracle's out-of-sample accuracy is low -- an unavoidable reality in financial forecasting. On three real-world crises, \MethodName improves scenario diversity, realism, and tail-risk coverage over RL and MCMC baselines, and ablations over oracle architectures and distribution families yield practical insights into continuous GFlowNet behavior under noisy rewards.
\end{abstract}


\section{Introduction}
\label{sec:introduction}

Effective financial risk management hinges upon anticipating tail risks -- rare, extreme events that result in disproportionately large losses. Traditional techniques, including historical simulation and Value-at-Risk (VaR), often fall short in modeling such events, relying on historical data and well-behaved distributional assumptions that underestimate the likelihood and severity of extreme outcomes~\cite{jorion2006var}.

Recent advancements have sought to overcome these limitations. Sampling-based methods like Sequential Markov Chain Monte Carlo (SMCMC) produce diverse scenarios but cannot efficiently target high-risk regions. Deep Reinforcement Learning (DRL) approaches discover high-reward strategies but converge to narrow behavioral sets, sacrificing diversity.

This paper introduces \MethodName, which applies Generative Flow Networks (GFlowNets) to tail-risk scenario discovery. GFlowNets learn policies that sample trajectories proportionally to a reward function, enabling principled exploration of high-impact outcomes~\cite{bengio2021flow}. By designing rewards that prioritize adverse financial states, our framework generates diverse, realistic, high-severity scenarios that existing methods fail to capture.

A key practical advantage is robustness to imperfect reward signals. Unlike RL methods that overfit to reward noise, GFlowNets' proportional sampling maintains broad exploration even when the oracle has limited accuracy -- a common reality in financial forecasting, where out-of-sample $R^2$ values are typically low or negative~\cite{campbell2008predicting}. We empirically validate that our approach generates realistic crisis trajectories despite oracle limitations (\cref{subsec:TrajectoryRealism}).

Importantly, the problem \MethodName addresses is not specific to finance. The defining characteristics -- continuous, interdependent variables; sequential temporal dependencies; a noisy, imperfect reward proxy; and a need for \emph{diverse} rather than single optimal solutions -- are domain-agnostic, and the same formulation transfers to settings such as power-grid resilience and supply-chain stress testing (\cref{sec:conclusions}). The financial indicators (VIX, credit spreads, equity indices) are concrete instantiations of these abstract properties. Our contributions are:

\begin{itemize}
    \item \textbf{Continuous GFlowNets for real-world sequential risk discovery.} To our knowledge, \MethodName is the first deployment of continuous GFlowNets~\cite{lahlou2023theory} on a real-world, high-dimensional sequential decision problem with an imperfect oracle, instantiated for macro-financial tail-risk discovery.
    \item \textbf{A flexible mixture-policy architecture.} We introduce a policy that combines \emph{variable-specific} distribution families within a single network via a Gumbel-Softmax mixture, enabling per-variable, feature-specific heteroscedasticity -- an architectural design not addressed by prior continuous GFlowNet work.
    \item \textbf{Trajectory-level diversity evaluation.} We propose a Dynamic-Time-Warping (DTW) diversity metric that captures full temporal structure, complementing terminal-state metrics. It is a general-purpose tool for any sequential generative model.
    \item \textbf{Comprehensive empirical study.} On a newly constructed 1996--2023 macro-financial dataset spanning three crises, \MethodName outperforms RL and MCMC baselines on diversity, realism, and tail-risk coverage. Ablations over oracle architectures and distribution families provide practical insights into continuous GFlowNet behavior under noisy rewards.
\end{itemize}
We emphasize that several individual components (continuous GFlowNets, time-indexing, replay buffers) originate in prior work; our contribution lies in their integration, the mixture-policy architecture, the domain adaptation, the evaluation methodology, and the empirical insights.

\section{Related Work}
\label{sec:RelatedWork}

We review methodologies relevant to financial risk scenario modeling, grouped into deep generative models, reinforcement learning, and probabilistic simulation.

\textbf{Deep Generative Models} such as GANs~\cite{zhang2018generative}, VAEs~\cite{desai2021timevae}, diffusion models~\cite{yuan2024diffusion}, and LLMs~\cite{cao2024risklabs,yuksel2025alphasharpe} are increasingly used in finance for synthetic market data generation and anomaly detection. While GANs generate realistic sequences, they often suffer from mode collapse and are not designed to model rare or tail-risk scenarios, limiting their applicability in high-diversity risk discovery.

\textbf{Reinforcement Learning (RL)} has been widely applied to portfolio optimization~\cite{cui2024multi} and risk management~\cite{anwar2024deep}. Traditional RL methods (DQN, policy gradients~\cite{sutton1999policy}, actor-critic~\cite{konda1999actor}) optimize \emph{expected} returns, leading agents to converge on a small set of strategies and hindering exploration of diverse failure modes. RL's reward-maximization objective is also particularly vulnerable to oracle misspecification: agents overfit to spurious high-reward predictions rather than exploring diverse failure modes. We note that GFlowNets can themselves be viewed as a form of entropy-regularized RL; formal equivalences have been established in \emph{discrete} settings~\cite{tiapkin2024generative,deleu2024discrete}. Our comparison therefore contrasts trajectory-balance training against a strong entropy-regularized RL baseline (Soft Actor-Critic); empirically, this baseline still collapses to narrower scenario sets in our continuous, sequential domain (\cref{sec:experiments}).

\textbf{Probabilistic Simulation Methods} such as Sequential MCMC (SMCMC)~\cite{andrieu1999sequential}, jump-diffusion models~\cite{rachev2003handbook}, and Bayesian Optimization (BO)~\cite{shahriari2015taking} offer interpretable, uncertainty-aware alternatives. SMCMC produces diverse trajectories but incurs high computational cost and slow convergence in high dimensions. Jump-diffusion models embed rare events into stochastic processes for stress testing but require strong parametric assumptions. BO identifies extreme outcomes with few evaluations~\cite{frazier2018tutorial,cheon2024earl} but typically targets global optima, limiting trajectory diversity.

\textbf{Generative Flow Networks} have been applied to molecule and biological-sequence design~\cite{bengio2021flow, jain2022biological}, combinatorial scheduling~\cite{zhang2023robustschedulinggflownets}, and continuous control~\cite{li2023cflownetscontinuouscontrolgenerative}, with active research on training stability and credit assignment~\cite{malkin2023trajectorybalanceimprovedcredit, shen2023towards, kim2024logitgfn} and distributional extensions~\cite{zhang2024distributionalgflownetsquantileflows}. We adapt this family to sequential financial scenario generation; the closest application to finance, AlphaSAGE~\cite{chen2025alphasage}, mines predictive alpha factors and is complementary to our tail-risk discovery setting.

\begin{figure*}[t]
    \centering
    \includegraphics[width=0.9\linewidth]{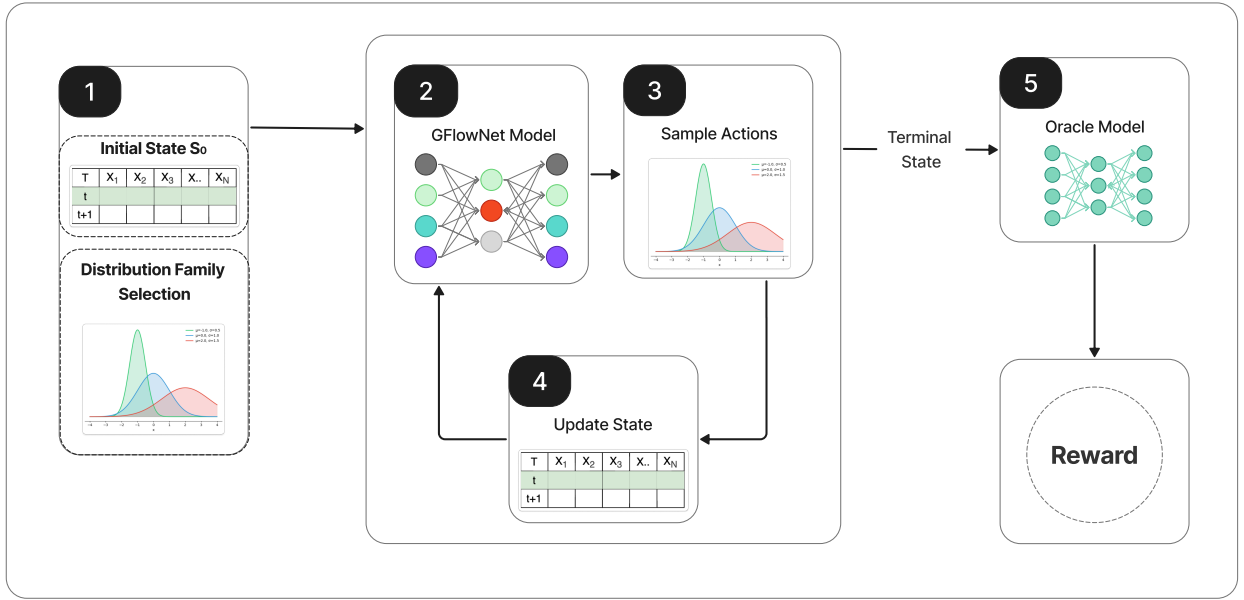}
    \caption{An overview of \MethodName. The method begins by modeling each environmental variable and the initial state, then explores the state space by sampling trajectories. Sampling is guided by the GFlowNet loss and feedback from the oracle.}
    \label{fig:methodology}
\end{figure*}

\section{Background}
\label{sec:Background}

GFlowNets learn a generalized stochastic policy~\cite{atanackovic2024investigating} that constructs complex objects through sequential actions, with the goal of sampling each object $x$ with probability proportional to a non-negative reward $R(x)$, rather than maximizing cumulative reward as in standard RL~\cite{bengio2021flow}. By framing generation as a flow network over a directed acyclic graph of partial states, GFlowNets convert an unnormalized target distribution into an amortized sampler that produces independent, diverse samples in one shot, avoiding the slow mixing and mode collapse of MCMC and the limited diversity of deep RL~\cite{bengio2023gflownet}.

At the core of a GFlowNet is a flow function $F$ over edges $(s \to s')$ that must satisfy a conservation constraint at every intermediate state,
\begin{equation}
    \sum_{s': (s,s') \in \mathcal{E}} F(s \to s') = \sum_{s'': (s'',s) \in \mathcal{E}} F(s'' \to s),
    \label{eq:flow_conservation}
\end{equation}
ensuring total inflow equals total outflow except at the initial and terminal states~\cite{bengio2021flow}. Flows to terminal states equal their rewards, and the forward policy is
\begin{equation}
    \pi(s' \mid s) = \frac{F(s \to s')}{F(s)}, \quad F(s) = \!\!\sum_{s': (s,s') \in \mathcal{E}}\!\! F(s \to s'),
    \label{eq:forward_policy}
\end{equation}
which guarantees that the induced marginal probability of generating a terminal state $x$ is $P_F(x) = R(x) / \sum_{x'} R(x')$. This alignment between policy and reward lets GFlowNets sample high-reward scenarios more frequently while retaining support over all modes~\cite{bengio2023gflownet}. Three properties make them effective for financial risk: \emph{amortized one-shot sampling} (efficient generation without iterative procedures), \emph{reward-aligned exploration} (mass directed toward high-impact scenarios, mitigating RL-style mode collapse), and \emph{multi-modal modeling} (capturing distinct failure modes, e.g., liquidity crises vs.\ inflation shocks, simultaneously).

Recent work extends GFlowNets to continuous, high-dimensional domains~\cite{lahlou2023theory, brunswic2024theory} using support measures and transition kernels, and introduces training objectives -- Detailed, Trajectory, and Sub-Trajectory Balance~\cite{malkin2023trajectorybalanceimprovedcredit, madan2023learninggflownetspartialepisodes} -- that enforce flow consistency at different granularities. We adopt Trajectory Balance for its strong credit assignment over full trajectories~\cite{malkin2023trajectorybalanceimprovedcredit}, which suits our fixed-horizon sequential setting. Most such applications concern single-shot sampling from a \emph{known} density (e.g., diffusion-structured samplers~\cite{sendera2024improved}) and are validated on low-dimensional synthetic settings such as 2D grids, 1D control, or 2D tori~\cite{lahlou2023theory}. \MethodName instead targets real-world, high-dimensional sequential generation (6 variables over 12 timesteps) with temporal dependencies and an imperfect, \emph{learned} oracle.

\section{Problem Formulation}
\label{sec:problem}

We formalize tail-risk scenario discovery as generation over a finite-horizon environment. A state $s_t = (v_t, t)$ comprises $d$ macro-financial variables $v_t \in \mathbb{R}^d$ (\cref{tab:initial_state}) and a discrete time index $t \in \{0, \dots, T\}$. The initial state $s_0$ fixes a pre-crisis macroeconomic condition. An action $a_t \in \mathbb{R}^d$ specifies the change applied to each variable, and the transition
\begin{equation}
    s_{t+1} = f(s_t, a_t)
    \label{eq:transition}
\end{equation}
applies $a_t$ under affine and domain-clipping constraints (\cref{sec:transitions}). Trajectories have a fixed horizon $T$ ($T=12$ months), defining termination. A terminal trajectory $\tau = (s_0, \dots, s_T)$ receives reward
\begin{equation}
    R(s_T) = \max\!\big(0,\, -\mathcal{O}(s_T)\big),
    \label{eq:reward_function}
\end{equation}
where $\mathcal{O}$ is a pre-trained return-forecasting oracle (\cref{subsec:reward}).

Crucially, unlike standard RL -- which seeks a policy maximizing $\mathbb{E}[R]$ -- we seek a sampler whose terminal distribution is \emph{proportional} to reward, $P(\tau) \propto R(s_T)$, so that many diverse high-risk scenarios are discovered rather than a single optimum. The variables are continuous and interdependent, the horizon is sequential with temporal dependencies, and the oracle is an imperfect proxy. These problem characteristics -- not the financial labels -- determine the method, which is why the same formulation applies to other domains with continuous interdependent variables and noisy oracles.

\section{The \MethodName Method}
\label{sec:method}

\textbf{Overview.} We present our \textbf{G}FlowNet-based \textbf{RI}sk \textbf{D}etection (\MethodName) method in \cref{fig:methodology}. We specify the initial macroeconomic state, a parametric distribution family per environmental variable (learned by the network), and a fixed horizon (step \#1). At each step the GFlowNet predicts the action-distribution parameters (step \#2), samples a transition (step \#3), and updates the state (step \#4). Upon reaching the terminal state, a predictive oracle evaluates the trajectory and returns a scalar reward used to train the policy via flow-matching objectives (step \#5).

The architecture has three advantages over existing scenario generators. First, it is strongly explorative, particularly with flexible families such as mixtures of Gaussian or Beta functions. Second, it supports a \emph{different} distribution function per variable, enabling accurate environment modeling and edge-case exploration -- a capability that energy- or diffusion-based methods do not readily provide. Third, once trained the policy is an amortized sampler, generating high-reward scenarios without iterative inference, unlike SMCMC or optimization-based approaches.

\begin{algorithm}[t]
\caption{\MethodName: GFlowNet-based Risk Discovery}
\label{alg:grid}
\begin{algorithmic}[1]
\Require initial state $s_0$, horizon $T$, per-variable families $\{\mathcal{F}_i\}$, oracle $\mathcal{O}$, replay buffer $\mathcal{B}$
\State Initialize forward/backward policies $\pi_F, \pi_B$ and $\log Z$
\While{not converged}
  \State $s \gets s_0$; initialize trajectory with $s_0$
  \For{$t = 0$ to $T-1$}
    \State Predict mixture parameters $(w, \mu, \sigma)$ from $\pi_F(\cdot \mid s)$
    \State $\alpha \gets \mathrm{GumbelSoftmax}(w, \lambda)$ \Comment{Eq.~\eqref{eq:gumbel_softmax}}
    \State Sample action $a$ from the mixture \Comment{Eq.~\eqref{eq:mixture_distribution}}
    \State $s \gets f(s, a)$ with plausibility clipping \Comment{Eq.~\eqref{eq:transition}}
    \State Append $s$ to the trajectory
  \EndFor
  \State $R \gets \max(0, -\mathcal{O}(s))$ \Comment{terminal reward, Eq.~\eqref{eq:reward_function}}
  \State Store the trajectory in $\mathcal{B}$; sample a minibatch from $\mathcal{B}$
  \State Update $\pi_F, \pi_B, \log Z$ via $\mathcal{L}_{\text{TB}}$ \Comment{Eq.~\eqref{eq:trajectory_balance_loss}}
\EndWhile
\State \Return amortized sampler $\pi_F$
\end{algorithmic}
\end{algorithm}

\subsection{State and Environment-Variable Distributions}
\label{subsec:distributions}

\textbf{The initial state} begins at a specified macroeconomic state, comprising core indicators such as equity indices and trading volumes plus a timestep $t$ (\cref{tab:initial_state}).

\begin{table}[t]
\centering
\caption{Initial macroeconomic environment variables defining the starting state $s_0$.}
\label{tab:initial_state}
\footnotesize
\begin{tabular}{ll}
\toprule
\textbf{Variable} & \textbf{Description} \\
\midrule
Volume\_spx & Trading volume of the S\&P 500 index \\
Close\_ndx & Closing price of the NASDAQ-100 index \\
Volume\_ndx & Trading volume of the NASDAQ-100 index \\
Close\_vix & Closing value of the VIX (volatility index) \\
IRLTCT01USM156N & Long-term US interest rate \\
BAMLH0A3HYCEY & High-yield corporate bond spread \\
\bottomrule
\end{tabular}
\end{table}

\textbf{Modeling variable distributions.} While continuous GFlowNets permit arbitrary transition kernels in theory~\cite{lahlou2023theory}, they do not address combining \emph{heterogeneous} families within a single policy network. \MethodName does so through a three-step procedure:

\emph{(1) Domain-guided family assignment.} Each variable is assigned a family from its statistical properties: Beta for bounded, asymmetric variables (e.g., VIX), Gaussian mixtures for interest rates, and Student-t for heavy-tailed volumes. We support Gaussian, Gaussian Mixture, Beta, Mixture Beta, and Student-t.

\emph{(2) Policy parameterization.} At each step the policy outputs the family's parameters. For a $K$-component mixture it predicts logits $w \in \mathbb{R}^K$, means $\mu \in \mathbb{R}^K$, and scales $\sigma \in \mathbb{R}^K$. Mixture weights use a temperature-controlled Gumbel-Softmax~\cite{jang2016categorical},
\begin{equation}
    \alpha = \mathrm{GumbelSoftmax}(w, \lambda),
    \label{eq:gumbel_softmax}
\end{equation}
with learnable or fixed temperature $\lambda$, and the action is drawn from the weighted combination
\begin{equation}
    a \sim \mathcal{N}\!\left(\textstyle\sum_{k=1}^K \alpha_k \mu_k,\; \sum_{k=1}^K \alpha_k \sigma_k \right),
    \label{eq:mixture_distribution}
\end{equation}
producing a unimodal approximation that retains multimodal expressivity during training and supports feature-specific heteroscedasticity.

\emph{(3) Learning.} Parameters are refined by the trajectory-balance loss (\cref{subsec:training}) toward high-risk transitions, while the fixed family preserves appropriate support and tail behavior. Thus when extreme, higher-reward values are encountered, the learned distribution assigns those edge cases greater probability of being explored.

\subsection{State Transitions and Time-Indexing}
\label{sec:transitions}

At each timestep, the sampled action -- a distribution over economic-variable changes (credit spreads, interest rates, equity indices, etc.) -- is applied to the current state to produce the next, as in \cref{eq:transition}, under affine transformation and domain-clipping constraints.

\begin{figure}[t]
    \centering
    \includegraphics[width=\linewidth]{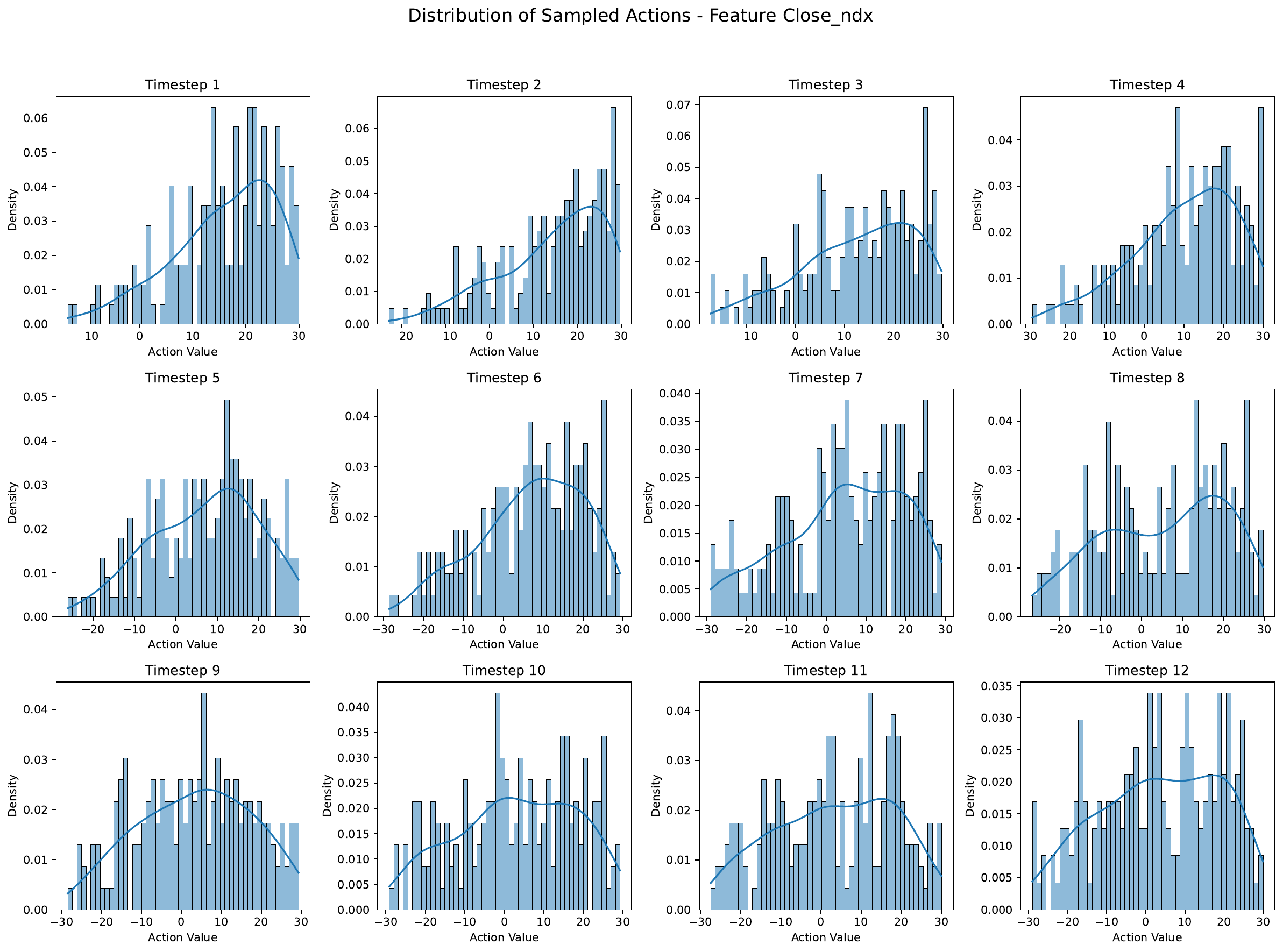}
    \caption{Learned action distribution for the \texttt{Close\_ndx} variable at each of the 12 timesteps. The distribution shifts markedly across the horizon, showing that the time index enables \MethodName to learn time-dependent dynamics rather than applying a stationary policy.}
    \label{fig:time_evolution}
\end{figure}

We incorporate a \emph{time index} into the state representation, following continuous GFlowNet formulations~\cite{lahlou2023theory}. Beyond preventing degenerate cycles -- since $(s_t, t) \ne (s_{t'}, t')$ for $t \ne t'$ -- the time index lets the policy learn \emph{time-dependent} action distributions that evolve across the horizon (\cref{fig:time_evolution}), capturing how crisis dynamics shift over time.

To keep transitions plausible, we bound per-step changes using historical data: while we want diverse scenarios, changes must remain reasonable (e.g., not a 20\% interest-rate move in one month). We sampled a large number of sequential steps from the real-world training set and recorded the maximal single-step mean and standard deviation; \MethodName is then prevented from generating fluctuations larger than these. Because the mean and standard deviation need not come from the same sequence, the model retains \emph{greater} flexibility than any single historical episode while remaining bounded.

\subsection{Terminal-State Evaluation, Reward, and Oracle Imperfection}
\label{subsec:reward}

Once a trajectory reaches its terminal state $s_T$, a pre-trained oracle predicts a downstream outcome -- the projected 12-month percentage return of the S\&P~500 -- mapped to a scalar reward via \cref{eq:reward_function}. This assigns zero value to positive market movements while positively rewarding adverse evolutions. We train with a single MLP oracle and evaluate with a separate model ensemble (\cref{subsec:ExperimentalSetup}) so that scenarios are not artifacts of one model's inductive biases.

\textbf{Oracle imperfection as a design assumption.} Financial return prediction is inherently noisy: out-of-sample $R^2$ in equity-return forecasting is routinely near zero or negative, yet such predictors carry statistically significant, economically meaningful signal~\cite{campbell2008predicting, hou2013r}. GFlowNets are well-suited to this regime because they sample proportionally to reward rather than maximizing it, spreading mass across multiple high-reward modes instead of over-exploiting noise-driven peaks~\cite{bengio2021flow, jain2022biological}. This is a core methodological advantage, not a limitation to overcome, and we validate it empirically (\cref{subsec:TrajectoryRealism,subsec:ablations}).

\textbf{Why \MethodName's scenarios are realistic.} Three design choices drive realism. (i) The historical-plausibility constraints (\cref{sec:transitions}) bound per-step changes to observed magnitudes, ruling out implausible trajectories. (ii) Proportional sampling avoids the anti-realistic trajectories that pure reward maximization tends to produce: as \cref{tab:trajectory_realism} shows, RL baselines correlate \emph{negatively} with real crises in several cases, having exploited oracle noise. (iii) Variable-specific families keep bounded variables bounded and heavy-tailed variables appropriately heavy-tailed.

\subsection{Training Objective and Replay Buffer}
\label{subsec:training}

The GFlowNet is optimized using the \emph{trajectory balance} loss, which enforces that the likelihood of sampling a trajectory is proportional to its final reward:
\begin{equation}
\begin{split}
    \mathcal{L}_{\text{TB}} = \Big( &\log Z + \textstyle\sum_{t=0}^{T-1} \log \pi_F(s_{t+1} \mid s_t) \\
    &- \textstyle\sum_{t=1}^{T} \log \pi_B(s_{t-1} \mid s_t) - \log R(s_T) \Big)^{2},
\end{split}
    \label{eq:trajectory_balance_loss}
\end{equation}
where $\pi_F$ and $\pi_B$ are the forward and backward policies and $Z$ is the learnable partition function. To enhance sample efficiency and stability, we incorporate a replay buffer~\cite{mnih2013playing,shen2023towards,vemgal2023empirical} -- a standard component in GFlowNet training -- that stores past trajectories for resampling, smoothing gradient estimates. In continuous spaces there is no natural fixed backward policy (unlike the uniform-over-parents choice common in discrete GFlowNets), so we train $\pi_B$ jointly; we examine this choice in \cref{subsec:ablations}.

\section{Evaluation \& Analysis}
\label{sec:experiments}

Our evaluation answers four questions. \textbf{(RQ1)} Does \MethodName discover more \emph{severe} tail-risk scenarios than existing methods? \textbf{(RQ2)} Does it do so while preserving scenario \emph{diversity}? \textbf{(RQ3)} Are the generated scenarios \emph{realistic} despite a noisy, low-$R^2$ oracle? \textbf{(RQ4)} How sensitive is performance to the oracle and distribution-family choices? We address RQ1 via reward and tail-risk metrics (\cref{subsec:results,subsec:TailRiskMetrics}), RQ2 via diversity metrics (\cref{subsec:RiskScenarioDiversity}), RQ3 via trajectory realism (\cref{subsec:TrajectoryRealism}), and RQ4 via ablations (\cref{subsec:ablations}).

\subsection{Datasets and Baselines}
\label{sec:baselines}

\textbf{Dataset.} We constructed a macro-financial dataset from public sources (FRED\footnote{\url{https://fred.stlouisfed.org/}}, Yahoo Finance) covering 1996--2023, extracting the variables in \cref{tab:initial_state} at monthly intervals, normalized to $[-100, 100]$. The 27-year span contains three major crises that we use as test sets: the Dot-com bubble (2001--2002), the Global Financial Crisis (2007--2008), and the COVID-19 crash (2020--2022).

\textbf{Baselines.} We compare \MethodName to three established scenario-discovery baselines covering the main exploration paradigms (\cref{sec:RelatedWork}): SMCMC, REINFORCE, and Soft Actor-Critic (SAC). \emph{All methods share the same horizon, oracle reward, and action space.}

\textbf{1) \MethodName} -- trained with the trajectory-balance loss; forward and backward policies are 3-layer MLPs (LeakyReLU, layer norm). Trajectories span 12 timesteps; training uses a replay buffer of size 5000, Adam (lr $5\times10^{-4}$) for both networks and $\log Z$, and batch size 16. We sample 200 trajectories for evaluation.

\textbf{2) SMCMC} -- Gaussian proposal (std 10, Boltzmann temperature 10); fully offline. For each use case we generate 2{,}000 trajectories and retain the top 200 by oracle reward, for fair comparison with the learned methods.

\textbf{3) REINFORCE}~\cite{williams1992simple} -- a 3-layer MLP actor (hidden dim 128, LeakyReLU) predicting Gaussian-policy parameters, trained with trajectory-level rewards for 1{,}500 episodes; 200 trajectories for evaluation.

\textbf{4) SAC}~\cite{haarnoja2018soft} -- a Gaussian policy with twin Q-networks (policy 2-layer MLP dim 64; critics 2-layer MLP dim 128), target updates $\tau=0.0005$, auto-tuned entropy temperature ($\alpha_0=0.5$), discount $\gamma=0.999$, trained for 1{,}500 episodes; 200 trajectories for evaluation.

We tuned multiple configurations and training times per baseline and report the best; additional training did not improve baseline performance.

\subsection{Experimental Setup}
\label{subsec:ExperimentalSetup}

\textbf{Oracle pretraining.} The oracle estimates future market returns from historical macro-financial data; all experiments were run on a commodity laptop CPU. To eliminate data leakage, the oracle is trained exclusively on pre-crisis data per scenario (e.g., for the Dot-com case, 1996--2000, evaluated on trajectories beginning in 2001), guaranteeing the GFlowNet explores strictly forward in time. The training-phase oracle is a lightweight 3-layer MLP (32 hidden units/layer, LeakyReLU, layer norm), trained with Adam (lr $10^{-4}$) for 300 epochs to minimize MSE. Oracle metrics are in \cref{tab:oracle_metrics}.

\begin{table}[t]
\centering
\setlength{\tabcolsep}{4pt}
\footnotesize
\caption{Oracle performance over a 24-month validation period. Lower MAE/RMSE is better; higher $R^2$ is better. Low or negative $R^2$ is typical in financial return prediction (see text).}
\label{tab:oracle_metrics}
\begin{tabular}{lccc}
\toprule
\textbf{Metric} & \textbf{Dot-Com (2002)} & \textbf{GFC (2008)} & \textbf{COVID-19 (2021)} \\
\midrule
$R^2 \uparrow$    & -0.4762 & 0.0942  & -0.8638 \\
RMSE $\downarrow$ & 11.59   & 4.2048  & 14.3532 \\
MAE $\downarrow$  & 9.78    & 2.92    & 12.81 \\
\bottomrule
\end{tabular}
\end{table}

\textbf{Interpreting oracle performance.} The $R^2$ values ($-0.86$ to $0.09$) must be read in context: low or negative out-of-sample $R^2$ is the empirical norm in equity-return forecasting, not a sign of failure. Seminal work shows that economically meaningful predictability coexists with near-zero or negative $R^2$~\cite{campbell2008predicting}, and that $R^2$ measures unconditional variance explanation rather than information content~\cite{hou2013r, kelly2013market}. Our oracle's non-trivial MAE (2.92--12.81) indicates it still captures the level and sign of returns. We further validate oracle utility through trajectory realism (\cref{subsec:TrajectoryRealism}): despite low $R^2$, \MethodName generates scenarios highly correlated with actual crises (\cref{tab:trajectory_realism}).

\textbf{Oracle models.} We intentionally train with a single MLP oracle but evaluate with a four-model ensemble (Random Forest, Gradient Boosting, Elastic Linear Model, MLP), reflecting practical finance settings where risk estimates are model-dependent and aggregated conservatively. This also serves as an implicit ablation: if the GFlowNet only learned the training oracle's boundaries, performance would degrade under the ensemble. \cref{tab:performance_metrics,tab:diversity_metrics} show strong performance and diversity under the ensemble, and the oracle-architecture ablation (\cref{tab:abl_oracle}) confirms \MethodName is not brittle to oracle choice, validating robustness.

\textbf{Evaluation metrics.} We report two reward metrics: (i) \emph{Median reward} -- the median oracle-predicted 12-month S\&P-500 change (higher is riskier; the median is robust to extreme values); and (ii) \emph{Q90 reward} -- the reward at the 90th percentile.

\textbf{Distribution selection.} We experimented with Gaussian, Student-t, and Beta families; Beta was affine-mapped into $[-30, 30]$ for bounded transitions. Mixtures markedly improved diversity and expressiveness (quantified in \cref{subsec:ablations}).

\begin{table}[t]
\centering
\caption{Key hyperparameters for \MethodName and the training-phase oracle. Baseline settings are given in \cref{sec:baselines}.}
\label{tab:hyperparams}
\footnotesize
\setlength{\tabcolsep}{5pt}
\begin{tabular}{ll}
\toprule
\textbf{Component / Setting} & \textbf{Value} \\
\midrule
\multicolumn{2}{l}{\emph{\MethodName{} policy ($\pi_F, \pi_B$)}} \\
\quad Architecture & 3-layer MLP, LeakyReLU, layer norm \\
\quad Horizon $T$ & 12 timesteps \\
\quad Replay buffer size & 5000 \\
\quad Optimizer / learning rate & Adam / $5\times10^{-4}$ \\
\quad Batch size & 16 \\
\quad Trajectories sampled (eval) & 200 \\
\quad Independent runs (seeds) & 30 \\
\midrule
\multicolumn{2}{l}{\emph{Training-phase oracle}} \\
\quad Architecture & 3-layer MLP, 32 units, LeakyReLU, LN \\
\quad Optimizer / learning rate & Adam / $10^{-4}$ \\
\quad Epochs / loss & 300 / MSE \\
\quad Evaluation oracle & ensemble (RF, GBR, ElasticNet, MLP) \\
\bottomrule
\end{tabular}
\end{table}

\subsection{Evaluation Results}
\label{subsec:results}

We performed 30 independent runs per algorithm, each with a different seed. For each run we trained a new model (except SMCMC, which is non-trainable) and sampled 200 trajectories from the final policy; for SMCMC we generated 2{,}000 per run and selected the top 200 by oracle reward.

\begin{table*}[t]
\centering
\caption{Performance metrics (Median and Q90 reward) over 30 runs per algorithm on three crises. Higher is better; rewards clipped at 100.}
\label{tab:performance_metrics}
\setlength{\tabcolsep}{6pt}
\small
\begin{tabular}{l|ccc|ccc}
\toprule
\textbf{Method} & \multicolumn{3}{c|}{\textbf{Median Reward}} & \multicolumn{3}{c}{\textbf{Q90 Reward}} \\
& 2002 & 2008 & 2021 & 2002 & 2008 & 2021 \\
\midrule
\MethodName & \cellcolor{highlight02}\textbf{43.11} & \cellcolor{highlight08}\textbf{31.13} & 18.12 & 58.57 & 54.52 & \cellcolor{highlight21}\textbf{100.00} \\
REINFORCE & 32.02 & 19.37 & 24.33 & \cellcolor{highlight02}\textbf{65.57} & 57.06 & 98.53 \\
SAC & 5.34 & 18.86 & 34.00 & 40.99 & \cellcolor{highlight08}\textbf{67.92} & \cellcolor{highlight21}\textbf{100.00} \\
SMCMC & 26.01 & 29.62 & \cellcolor{highlight21}\textbf{42.71} & 30.52 & 34.22 & 49.03 \\
\bottomrule
\end{tabular}
\end{table*}

\begin{table*}[t]
\centering
\caption{Diversity metrics across three crises: raw and reward-normalized scores for full trajectories (DTW-based) and terminal states. Higher is better. Results from 100 trajectories per method.}
\label{tab:diversity_metrics}
\scriptsize
\setlength{\tabcolsep}{4pt}
\resizebox{\textwidth}{!}{%
\begin{tabular}{l|ccc|ccc|ccc|ccc}
\toprule
\textbf{Method} & \multicolumn{3}{c|}{\textbf{Traj Div (DTW)}} & \multicolumn{3}{c|}{\textbf{Norm. Traj Div}} & \multicolumn{3}{c|}{\textbf{Terminal-State Div}} & \multicolumn{3}{c}{\textbf{Norm. Terminal-State Div}} \\
& 2002 & 2008 & 2021 & 2002 & 2008 & 2021 & 2002 & 2008 & 2021 & 2002 & 2008 & 2021 \\
\midrule
\MethodName & 108.85 & \cellcolor{highlight08}\textbf{550.39} & \cellcolor{highlight21}\textbf{399.15} & 99.02 & \cellcolor{highlight08}\textbf{481.41} & \cellcolor{highlight21}\textbf{399.15} & 79.80 & 266.13 & \cellcolor{highlight21}\textbf{200.81} & 72.60 & 232.78 & \cellcolor{highlight21}\textbf{200.81} \\
REINFORCE & 94.31 & 402.24 & 229.17 & 90.71 & 352.80 & 229.17 & 77.24 & \cellcolor{highlight08}\textbf{271.92} & 160.71 & 74.29 & \cellcolor{highlight08}\textbf{238.50} & 160.71 \\
SAC & \cellcolor{highlight02}\textbf{260.25} & 136.10 & 264.38 & \cellcolor{highlight02}\textbf{244.84} & 136.10 & 264.38 & \cellcolor{highlight02}\textbf{182.53} & 92.27 & 187.60 & \cellcolor{highlight02}\textbf{171.72} & 92.27 & 187.60 \\
SMCMC & 182.01 & 186.48 & 164.81 & 163.69 & 167.92 & 139.29 & 100.93 & 103.47 & 96.53 & 90.77 & 93.17 & 81.58 \\
\bottomrule
\end{tabular}%
}
\end{table*}

\begin{figure*}[t]
    \centering
    \begin{subfigure}{0.32\textwidth}
        \centering
        \includegraphics[width=\linewidth]{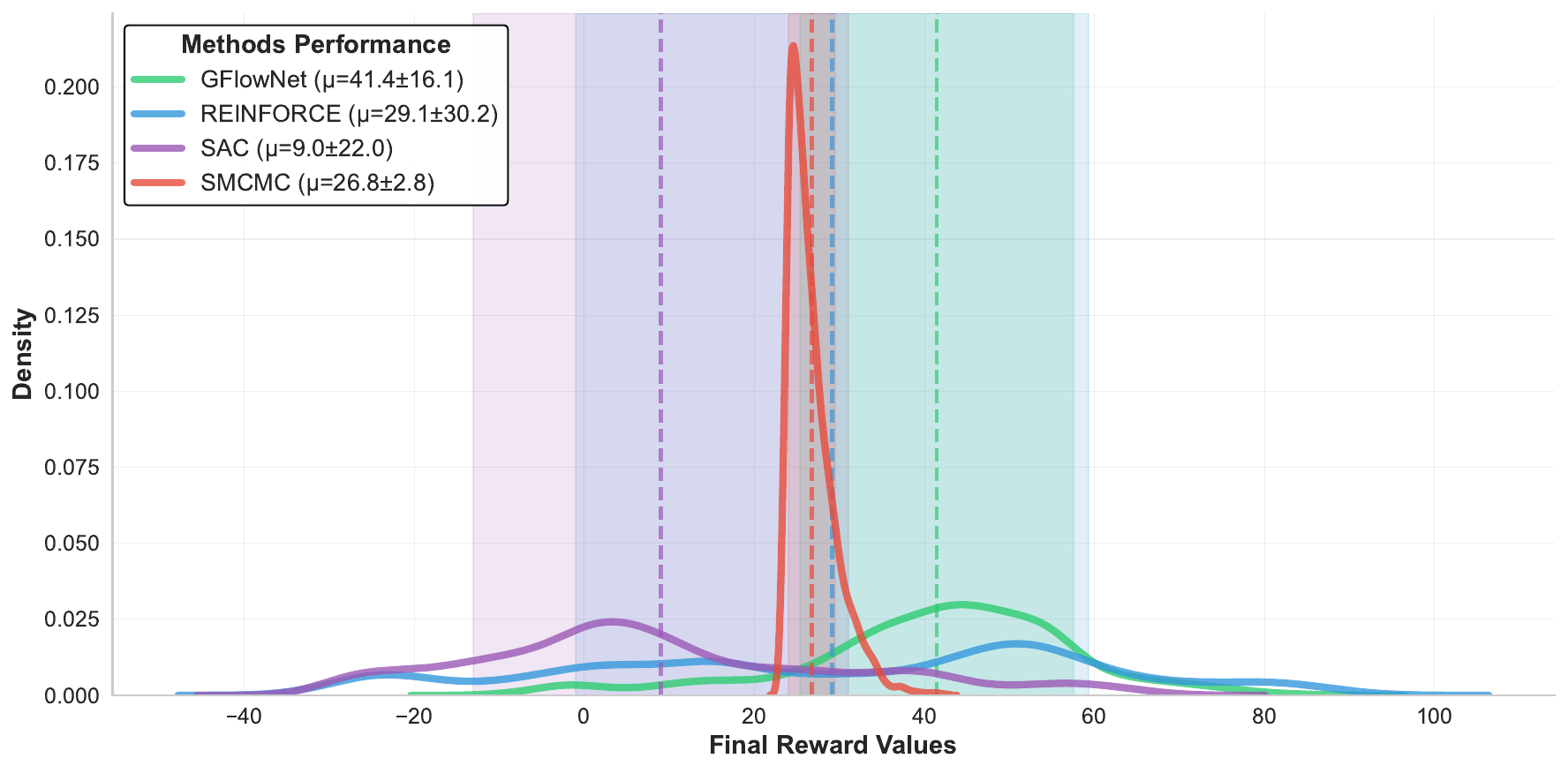}
        \caption{2002 (Dot-Com)}
        \label{fig:reward_dist_2002}
    \end{subfigure}\hfill
    \begin{subfigure}{0.32\textwidth}
        \centering
        \includegraphics[width=\linewidth]{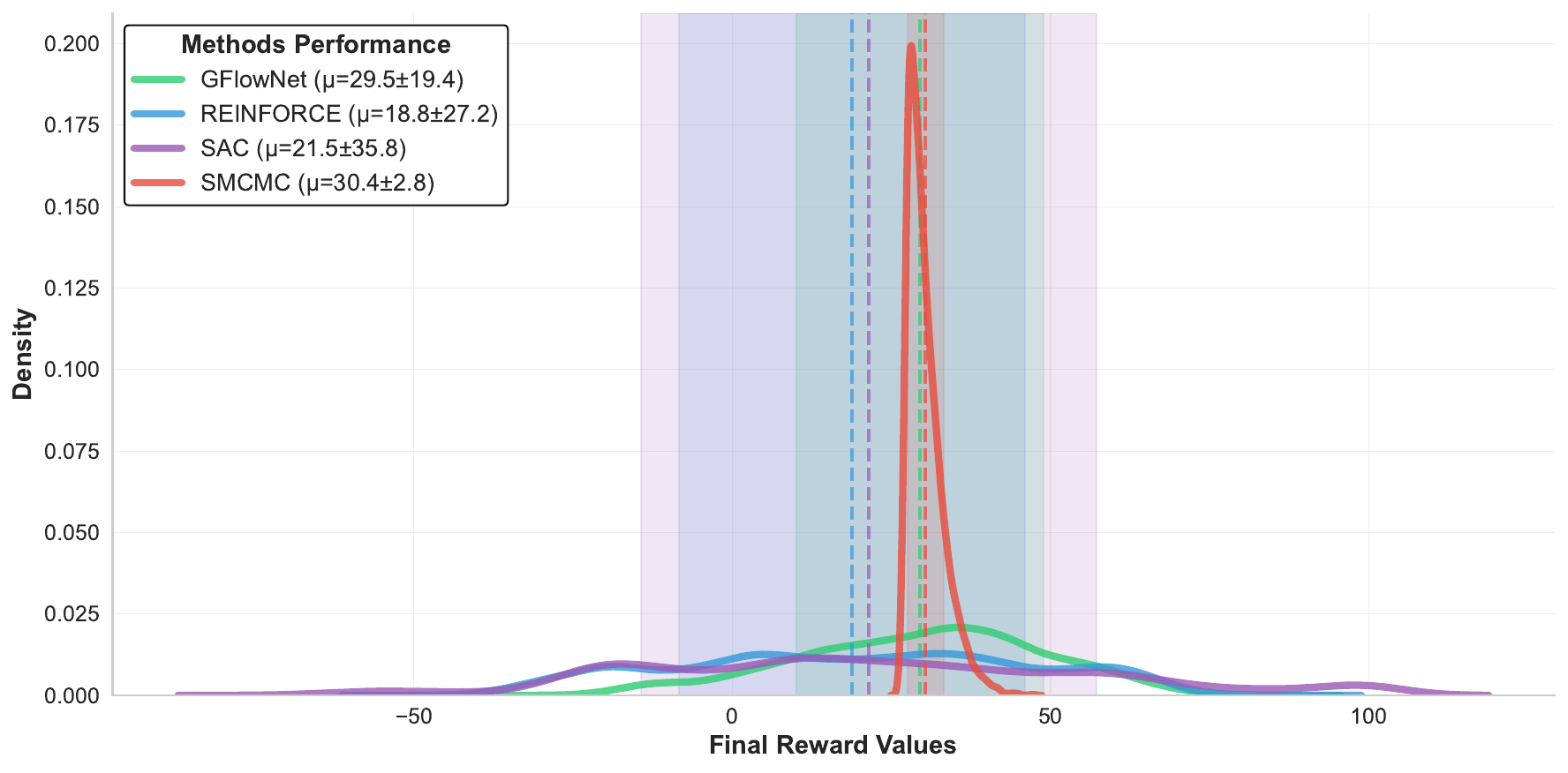}
        \caption{2008 (GFC)}
        \label{fig:reward_dist_2008}
    \end{subfigure}\hfill
    \begin{subfigure}{0.32\textwidth}
        \centering
        \includegraphics[width=\linewidth]{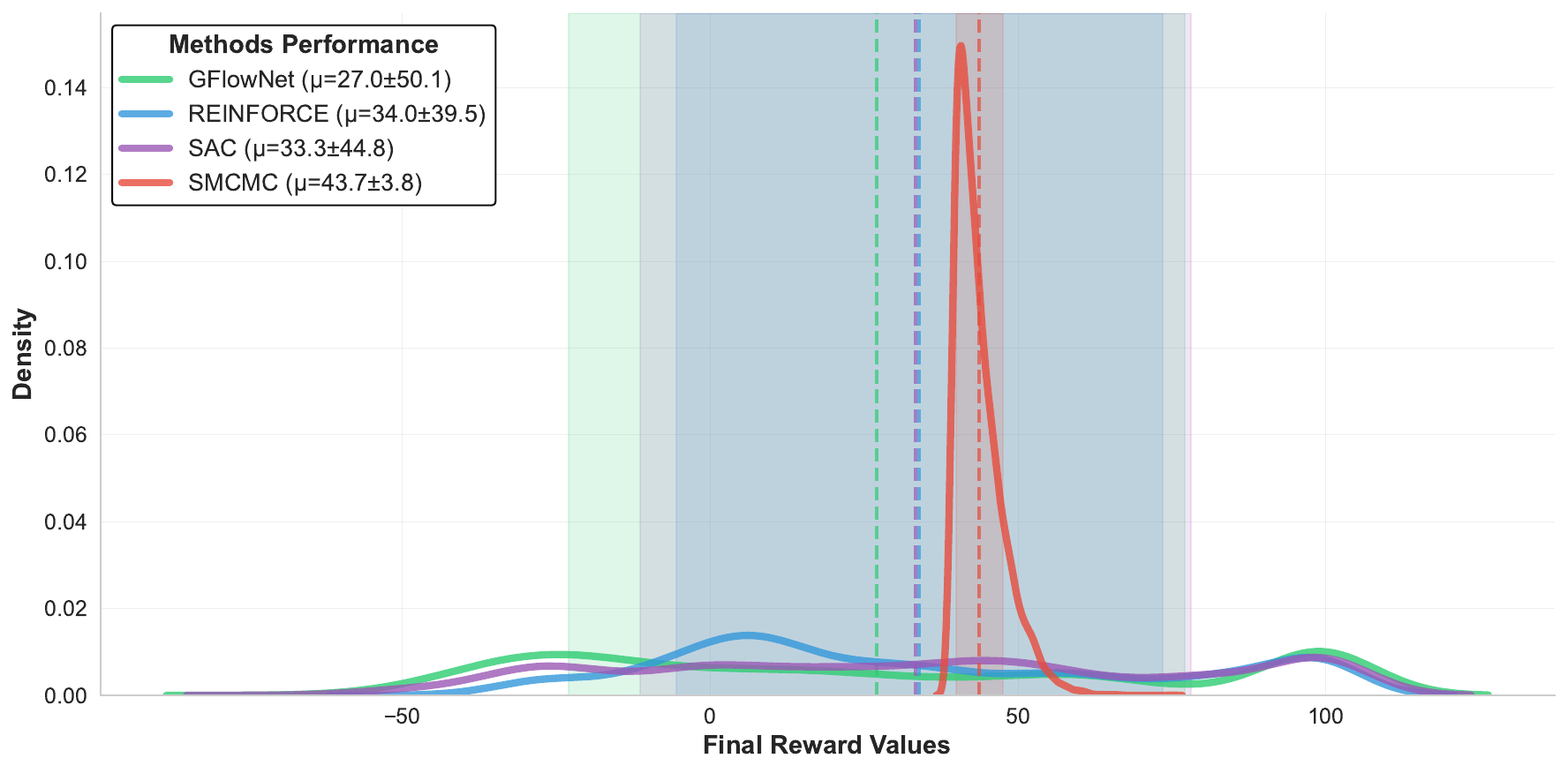}
        \caption{2021 (COVID-19)}
        \label{fig:reward_dist_2021}
    \end{subfigure}
    \caption{Reward distributions across 30 runs (6{,}000 trajectories per method) for the three crises. \MethodName produces a heavier high-reward tail and a broad spread, whereas SMCMC concentrates in a narrow band and the RL baselines place substantial mass on low rewards. The bimodality in 2021 (\subref{fig:reward_dist_2021}) explains \MethodName's lower median yet top Q90 there.}
    \label{fig:reward_dist_all}
\end{figure*}

\cref{tab:performance_metrics} shows \MethodName achieves the highest median reward in two of three crises. In 2021 the lower median reflects a \emph{bimodal} reward distribution (\cref{fig:reward_dist_2021}) -- two scenario clusters depress the median yet attain the top Q90. By contrast, SAC reaches Q90 67.92 in 2008 but median 18.86, a classic mode-collapse signature. In the Q90 metric \MethodName ties for best in 2021; while it is not top for Q90 in 2002/2008, its rewards are high and its medians far exceed those of SAC and REINFORCE, indicating it discovers \emph{many} high-reward scenarios rather than a few. Across all three crises (\cref{fig:reward_dist_all}), \MethodName exhibits a heavier high-reward tail than the baselines while avoiding SMCMC's narrow concentration, illustrating its superior diversity--severity trade-off.

\subsection{Analyzing Risk Scenario Diversity}
\label{subsec:RiskScenarioDiversity}

Existing GFlowNet evaluations typically focus on \emph{terminal-state} diversity. However, for sequential scenario generation the \emph{path} to the terminal state matters as much as the outcome -- different crisis pathways may demand different mitigation strategies even at similar final losses. We therefore introduce a trajectory-level diversity metric based on Dynamic Time Warping (DTW) that captures the full temporal structure, complementing terminal-state measures. This is a general-purpose tool applicable to any sequential generative model. We define:
\begin{equation}
    \text{Div}_{\text{Term}} = \frac{2}{N(N{-}1)} \sum_{i=1}^{N} \sum_{j=i+1}^{N} \big\lVert s_T^{(i)} - s_T^{(j)} \big\rVert_2,
    \label{eq:terminal_state_diversity}
\end{equation}
\begin{equation}
    \text{Div}_{\text{Traj}} = \frac{2}{N(N{-}1)} \sum_{i=1}^{N} \sum_{j=i+1}^{N} \text{DTW}\big(\tau^{(i)}, \tau^{(j)}\big),
    \label{eq:trajectory_diversity}
\end{equation}
with $\tau^{(i)} = (s_0^{(i)}, \dots, s_T^{(i)})$~\cite{sakoe2003dynamic}. We also report \emph{normalized} versions that factor in reward stability,
\begin{equation}
    \text{Norm} = \text{Div} \times \left(1 - \frac{|r_{\max} - r_{\min}|}{|r_{\max}| + |r_{\min}| + \epsilon}\right),
    \label{eq:reward_stability}
\end{equation}
penalizing methods that inflate diversity with many low-quality trajectories.

\MethodName attains top trajectory diversity for 2008 and 2021 (raw and normalized) and top terminal-state diversity for 2021, while remaining competitive elsewhere (\cref{tab:diversity_metrics}). SAC's and REINFORCE's high \emph{raw} diversity is largely an artifact of mode collapse: a few high-quality trajectories alongside many low-quality ones inflate pairwise distances without providing actionable coverage. \MethodName instead maintains high trajectory diversity while generating consistently high-reward populations, showing proportional sampling achieves both objectives at once.

\subsection{Trajectory Realism: Similarity to Actual Crises}
\label{subsec:TrajectoryRealism}

A natural concern is whether scenarios from a low-$R^2$ oracle resemble real crisis dynamics. We computed the Pearson correlation between each method's top-ranked trajectories (by risk score) and the actual macroeconomic changes during each crisis -- evaluating the temporal path, not just the endpoint.

\begin{table*}[t]
\centering
\caption{Trajectory similarity to actual crises: Pearson correlation between top-$K$ risk-ranked trajectories and observed macroeconomic dynamics. Higher is better; \textbf{bold} marks the best mean per crisis and $K$. \MethodName attains the best or competitive mean realism with consistently high diversity (std comparable to or exceeding the RL baselines), while the RL baselines frequently produce \emph{anti-realistic} (negatively correlated) trajectories.}
\label{tab:trajectory_realism}
\scriptsize
\setlength{\tabcolsep}{5pt}
\resizebox{\textwidth}{!}{%
\begin{tabular}{ll|ccc|ccc|ccc}
\toprule
& & \multicolumn{3}{c|}{\textbf{2002 (Dot-Com)}} & \multicolumn{3}{c|}{\textbf{2008 (GFC)}} & \multicolumn{3}{c}{\textbf{2021 (COVID-19)}} \\
\textbf{Method} & \textbf{Top-}$K$ & Max & Mean & Std & Max & Mean & Std & Max & Mean & Std \\
\midrule
\multirow{2}{*}{\MethodName} & 10 & 0.374 & \textbf{0.267} & 0.073 & 0.433 & 0.247 & 0.132 & 0.500 & \textbf{0.420} & 0.059 \\
 & 50 & 0.380 & \textbf{0.265} & 0.084 & 0.562 & 0.276 & 0.133 & 0.533 & \textbf{0.430} & 0.067 \\
\midrule
\multirow{2}{*}{REINFORCE} & 10 & 0.108 & 0.056 & 0.078 & -0.534 & -0.600 & 0.040 & 0.407 & 0.353 & 0.048 \\
 & 50 & 0.136 & 0.066 & 0.054 & -0.455 & -0.559 & 0.045 & 0.425 & 0.374 & 0.035 \\
\midrule
\multirow{2}{*}{SAC} & 10 & -0.179 & -0.243 & 0.044 & 0.416 & \textbf{0.374} & 0.024 & -0.124 & -0.143 & 0.013 \\
 & 50 & -0.161 & -0.247 & 0.047 & 0.452 & \textbf{0.377} & 0.026 & -0.114 & -0.144 & 0.019 \\
\midrule
\multirow{2}{*}{SMCMC} & 10 & 0.261 & 0.088 & 0.130 & 0.528 & 0.203 & 0.209 & 0.295 & 0.146 & 0.120 \\
 & 50 & 0.441 & 0.052 & 0.064 & 0.528 & 0.150 & 0.218 & 0.481 & 0.101 & 0.181 \\
\bottomrule
\end{tabular}%
}
\end{table*}

\MethodName achieves the highest \emph{mean} similarity in two of three crises (\cref{tab:trajectory_realism}), with maximum single-trajectory correlations of 0.56 (2008) and 0.53 (2021), and its mean is stable across top-10 and top-50, indicating that realism is not driven by a single lucky trajectory. It maintains substantial diversity (std comparable to or exceeding the RL baselines), suggesting it explores multiple plausible pathways rather than collapsing to one mode. The RL baselines instead frequently produce \emph{anti-realistic} trajectories -- SAC correlates negatively in 2002 and 2021, REINFORCE strongly so in 2008 -- because reward maximization exploits oracle noise. SMCMC shows the opposite failure: occasional high-max trajectories but low mean similarity, reflecting the absence of reward-proportional guidance. Overall, despite limited variance explanation, the oracle captures sufficient structure to guide \MethodName toward realistic scenarios when combined with proportional sampling.

\subsection{Risk Coverage with Standard Financial Metrics}
\label{subsec:TailRiskMetrics}

We further evaluate using standard stress-testing metrics: Maximum Drawdown (MaxDD, the greatest peak-to-trough loss), Value at Risk at 95\% (VaR$_{95}$, the 95th-percentile loss), and Conditional VaR at 95\% (CVaR$_{95}$, the expected loss beyond VaR$_{95}$).

\begin{table}[t]
\centering
\caption{Tail-risk metrics. Higher values indicate discovery of more severe scenarios. \MethodName achieves the highest severity metrics while maintaining diverse coverage.}
\label{tab:tail_risk_summary}
\scriptsize
\setlength{\tabcolsep}{3pt}
\begin{tabular}{l|ccc|ccc|ccc}
\toprule
& \multicolumn{3}{c|}{\textbf{MaxDD (\%)}} & \multicolumn{3}{c|}{\textbf{VaR$_{95}$ (\%)}} & \multicolumn{3}{c}{\textbf{CVaR$_{95}$ (\%)}} \\
\textbf{Method} & 2002 & 2008 & 2021 & 2002 & 2008 & 2021 & 2002 & 2008 & 2021 \\
\midrule
\MethodName & \textbf{34.3} & \textbf{100} & \textbf{100} & \textbf{30.4} & \textbf{94.0} & \textbf{90.5} & \textbf{22.2} & 38.3 & \textbf{30.3} \\
REINFORCE & 0.0 & 9.6 & 10.7 & 0.0 & 0.0 & 7.5 & 0.0 & 0.0 & 0.0 \\
SAC & 32.8 & 65.2 & 33.0 & 27.5 & 56.9 & 25.3 & 12.9 & \textbf{45.7} & 12.8 \\
SMCMC & 24.3 & 19.7 & 4.9 & 15.6 & 2.3 & 0.0 & 8.0 & 0.0 & 0.0 \\
\bottomrule
\end{tabular}
\end{table}

\MethodName achieves the highest MaxDD and VaR$_{95}$ in all scenarios (\cref{tab:tail_risk_summary}); its CVaR$_{95}$/VaR$_{95}$ ratios (0.33--0.73) indicate a \emph{spread} of severe scenarios rather than a single extreme. REINFORCE shows near-zero metrics (failure to explore tails despite the same oracle), SAC is moderate but inconsistent, and SMCMC, though diverse by design, fails to concentrate in high-severity regions.

\subsection{Ablation Studies}
\label{subsec:ablations}

We report four ablations addressing oracle sensitivity, distribution-family choice, the backward policy, and computational cost.

\textbf{Oracle architecture.} We trained \MethodName under five oracle architectures (\cref{tab:abl_oracle}). Three findings emerge. (i) \MethodName works with \emph{every} oracle, producing non-trivial reward and diversity in all 15 configurations -- it is not brittle to oracle choice. (ii) No single oracle dominates: Elastic Net leads reward in 2002 (81.0), MLP in 2008/2021 (80.2/77.7), and diversity leaders differ by scenario -- validating our train-MLP/evaluate-ensemble design and confirming \MethodName captures genuine crisis structure rather than one model's biases. (iii) \emph{Weaker oracles maintain or increase diversity}: GBR and RFR yield lower reward but higher normalized DTW diversity (e.g., 2021: GBR reaches the highest N.DTW, 216.2, at the lowest reward, 6.2). This is exactly the proportional-sampling behavior predicted by GFlowNet theory -- weaker reward signals broaden exploration rather than collapsing it -- empirically grounding the ``oracle imperfection as feature'' argument.

\begin{table}[t]
\centering
\caption{Oracle-architecture ablation: \MethodName trained under five oracles. Reward mean ($R_\mu$) and reward-normalized DTW diversity (N.DTW) per crisis.}
\label{tab:abl_oracle}
\scriptsize
\setlength{\tabcolsep}{4pt}
\begin{tabular}{l|cc|cc|cc}
\toprule
& \multicolumn{2}{c|}{\textbf{2002}} & \multicolumn{2}{c|}{\textbf{2008}} & \multicolumn{2}{c}{\textbf{2021}} \\
\textbf{Oracle} & $R_\mu$ & N.DTW & $R_\mu$ & N.DTW & $R_\mu$ & N.DTW \\
\midrule
Elastic Net & \textbf{81.0} & 98.3 & 24.9 & 165.5 & 38.6 & 83.2 \\
Ensemble & 41.4 & 97.4 & 29.4 & \textbf{496.3} & 29.8 & 83.6 \\
Gradient Boost. & 12.0 & 113.2 & 5.5 & 194.5 & 6.2 & \textbf{216.2} \\
MLP & 27.1 & 90.8 & \textbf{80.2} & 189.7 & \textbf{77.7} & 149.0 \\
Random Forest & 4.9 & \textbf{118.4} & 13.9 & 176.6 & 10.5 & 158.2 \\
\bottomrule
\end{tabular}
\end{table}

\textbf{Distribution family.} We compared three single families and two mixtures (\cref{tab:abl_dist}). (i) Family choice has \emph{dramatic} impact: among single families, reward spans 13.6, 85, and 73.7 points across the three crises -- an unsuitable family can cut performance 2--4$\times$. (ii) Beta consistently dominates, matching the bounded, asymmetric nature of financial variables; Cauchy yields high raw diversity but low reward and poor normalized diversity (scattered, unproductive exploration), validating domain-guided selection. (iii) Mixtures act as a \emph{safety net} (\cref{tab:abl_mixture}): a Gaussian mixture rescues a weak single Gaussian (+17--88\% reward, up to +485\% diversity) while roughly preserving a strong Beta ($-0.7\%$ to $-0.9\%$ reward). (iv) Mixtures improve diversity even without reward gains -- a Beta mixture raises N.DTW over single Beta in two of three crises (137.7$\to$171.8, 328.2$\to$410.3). For deployment where the optimal family is unknown a priori, this robustness justifies the Gumbel-Softmax mixture architecture (\cref{eq:gumbel_softmax,eq:mixture_distribution}).

\begin{table}[t]
\centering
\caption{Distribution-family ablation: reward mean ($R_\mu$) and reward-normalized DTW diversity (N.DTW) per crisis. ``Mix.'' marks mixture families.}
\label{tab:abl_dist}
\scriptsize
\setlength{\tabcolsep}{4pt}
\begin{tabular}{l|cc|cc|cc}
\toprule
& \multicolumn{2}{c|}{\textbf{2002}} & \multicolumn{2}{c|}{\textbf{2008}} & \multicolumn{2}{c}{\textbf{2021}} \\
\textbf{Family} & $R_\mu$ & N.DTW & $R_\mu$ & N.DTW & $R_\mu$ & N.DTW \\
\midrule
Beta & 35.7 & 137.7 & \textbf{100.0} & 181.0 & \textbf{100.0} & 328.2 \\
Cauchy & 22.2 & 81.4 & 15.0 & 106.1 & 30.4 & 153.4 \\
Beta (Mix.) & 35.4 & \textbf{171.8} & 76.4 & 128.1 & 99.3 & \textbf{410.3} \\
Gaussian (Mix.) & 34.5 & 139.6 & 75.0 & 134.7 & 49.3 & 48.4 \\
Gaussian & 29.4 & \textbf{213.3} & 56.0 & 98.9 & 26.3 & 61.1 \\
\bottomrule
\end{tabular}
\end{table}

\begin{table}[t]
\centering
\caption{Mixture effect: relative change ($\Delta\%$) of a mixture family over its corresponding single family (Gaussian mixture vs.\ Gaussian; Beta mixture vs.\ Beta), per crisis. Mixtures rescue weak families (large positive $\Delta$) while roughly preserving strong ones.}
\label{tab:abl_mixture}
\footnotesize
\setlength{\tabcolsep}{5pt}
\begin{tabular}{ll|ccc}
\toprule
\textbf{Crisis} & \textbf{Mixture} & $\Delta$Reward & $\Delta$Eucl. & $\Delta$DTW \\
\midrule
\multirow{2}{*}{2002} & Beta & $-0.9\%$ & $+23.3\%$ & $+11.2\%$ \\
 & Gaussian & $+17.3\%$ & $+22.8\%$ & $+65.1\%$ \\
\midrule
\multirow{2}{*}{2008} & Beta & $-23.5\%$ & $+3.9\%$ & $+8.3\%$ \\
 & Gaussian & $+33.8\%$ & $-7.7\%$ & $+22.1\%$ \\
\midrule
\multirow{2}{*}{2021} & Beta & $-0.7\%$ & $-5.5\%$ & $-6.5\%$ \\
 & Gaussian & $+87.6\%$ & $+485.1\%$ & $+194.1\%$ \\
\bottomrule
\end{tabular}
\end{table}

\textbf{Backward policy.} We additionally tried a \emph{fixed} Gaussian backward policy in place of the learned $\pi_B$. It exhibited poor convergence (notably for 2008 and 2021) and lower diversity due to constrained exploration, confirming that a learned backward policy improves credit assignment and stability in our continuous, sequential setting. Backward-policy choice in continuous GFlowNets remains understudied, and this result is a useful data point for the community.

\textbf{Computational cost.} \cref{tab:compute} reports cost on a commodity laptop CPU. \MethodName's training (23.3\,s) exceeds the RL baselines (3.9--7.2\,s), but the absolute cost is modest: end-to-end (training + 200 trajectories) is $\approx$44.6\,s, less than half of SMCMC's effective cost ($\approx$103.6\,s, since SMCMC must generate 10$\times$ more trajectories to reach comparable quality). Once trained, \MethodName generates each trajectory in $\approx$107\,ms, suitable for interactive stress testing -- a favorable profile given its diversity, realism, and tail-risk advantages.

\begin{table}[t]
\centering
\caption{Computational cost on a commodity laptop CPU. Inference is per generated trajectory. SMCMC is non-trainable but must generate 2{,}000 trajectories per use case (effective cost $\approx$103.6\,s).}
\label{tab:compute}
\footnotesize
\setlength{\tabcolsep}{8pt}
\begin{tabular}{lcc}
\toprule
\textbf{Method} & \textbf{Train (s)} & \textbf{Inference (ms/traj)} \\
\midrule
\MethodName & 23.3 & 106.8 \\
REINFORCE & 3.9 & 27.7 \\
SAC & 7.2 & 6.6 \\
SMCMC & -- & 51.8 \\
\bottomrule
\end{tabular}
\end{table}

\textbf{Computational complexity.} At each of the $T$ steps the policy predicts $O(dK)$ mixture parameters for $d$ variables and $K$ components, so generating a full trajectory costs $O(TdK)$ beyond the network forward pass and requires no iterative refinement -- the source of the amortized one-shot generation reflected in \cref{tab:compute}. Training adds a backward pass and a replay-buffer minibatch update per trajectory. The trajectory-level diversity evaluation (\cref{eq:trajectory_diversity}) costs $O(N^2 T d)$ for $N$ trajectories due to pairwise DTW; this dominates \emph{evaluation} but not generation and is computed offline.

\subsection{Discussion: Practical Implications and Insights}
\label{subsec:discussion}

\textbf{Practical implications for stress testing.} Beyond benchmark performance, \MethodName's profile fits operational risk workflows. Because the trained policy is an amortized sampler, it generates each scenario in $\approx$107\,ms (\cref{tab:compute}), enabling interactive, on-demand stress testing rather than the multi-hour reruns typical of simulation-based pipelines. Training with a single oracle while evaluating against a model ensemble mirrors standard practice, where risk estimates are model-dependent and aggregated conservatively, so scenarios that remain severe under the ensemble are more credible and transferable. Finally, because distinct failure modes may demand distinct mitigations, \MethodName's multi-modal, trajectory-level coverage -- as opposed to a single worst case -- is directly actionable for contingency planning.

\textbf{Insights transferable beyond finance.} Our study yields three findings of broader interest. First, it provides, to our knowledge, the first empirical evidence that continuous GFlowNets \emph{maintain or increase} sample diversity as the reward proxy weakens (\cref{tab:abl_oracle}): rather than collapsing, exploration broadens, confirming proportional-sampling theory in a genuinely noisy ($R^2<0$) regime relevant to any imperfect-oracle setting. Second, trajectory-balance training outperforms a strong entropy-regularized RL baseline (SAC) on diversity and realism in a continuous, sequential domain, complementing the discrete-space equivalences established in prior work~\cite{tiapkin2024generative,deleu2024discrete}. Third, the DTW-based trajectory-diversity metric (\cref{eq:trajectory_diversity}) is a general-purpose evaluation tool for any sequential generative model, addressing a gap in GFlowNet evaluations that focus on terminal states alone. \cref{tab:capability} summarizes the resulting trade-offs: no competing family simultaneously attains high tail severity, trajectory diversity, multi-modal coverage, and robustness to a noisy oracle, whereas \MethodName does.

\begin{table}[t]
\centering
\caption{Qualitative capability comparison, summarizing the trade-offs argued throughout and consistent with \cref{tab:performance_metrics,tab:diversity_metrics,tab:trajectory_realism,tab:tail_risk_summary}. \checkmark: supported; $\circ$: partial; $\times$: not supported; --: not applicable.}
\label{tab:capability}
\footnotesize
\setlength{\tabcolsep}{3pt}
\begin{tabular}{lcccc}
\toprule
\textbf{Capability} & \textbf{\MethodName} & \textbf{SMCMC} & \textbf{DRL} & \textbf{Deep Gen.} \\
\midrule
Amortized one-shot sampling & \checkmark & $\times$ & \checkmark & \checkmark \\
High tail severity (MaxDD/VaR) & \checkmark & $\times$ & $\circ$ & $\times$ \\
Trajectory diversity & \checkmark & \checkmark & $\times$ & $\circ$ \\
Multi-modal coverage & \checkmark & $\circ$ & $\times$ & $\times$ \\
Robust to noisy oracle & \checkmark & $\circ$ & $\times$ & -- \\
Historical plausibility constraints & \checkmark & $\times$ & $\times$ & $\times$ \\
\bottomrule
\end{tabular}
\end{table}

\section{Conclusion, Limitations, and Future Work}
\label{sec:conclusions}

We present \MethodName, a GFlowNet-based approach for discovering diverse financial tail-risk scenarios that outperforms RL and MCMC baselines on diversity, realism, and tail-risk coverage. To our knowledge, this is the first deployment of continuous GFlowNets on a real-world, high-dimensional sequential decision problem with an imperfect oracle, enabled by a flexible mixture-policy architecture, temporal mechanisms, and a trajectory-level evaluation methodology; our ablations yield practical insights into continuous GFlowNet behavior under noisy rewards.

Because \MethodName's core problem -- diverse trajectory generation over continuous, interdependent variables with a noisy oracle -- is domain-agnostic, the framework transfers directly to other high-impact settings such as power-grid resilience testing and supply-chain stress testing, with the oracle replaced by a domain simulator and the variable-specific families, temporal mechanisms, and trajectory-level diversity metric carried over unchanged.

\textbf{Limitations.} \MethodName's effectiveness depends on oracle quality, though we show that even low-$R^2$ oracles provide sufficient signal for realistic discovery, and that weaker oracles broaden (rather than degrade) exploration. Exploration is guided implicitly by the learned policy but can be steered through distribution families and mixture components.

\textbf{Threats to validity.} Our evaluation covers three historical crises; although these span distinct shock types (technology, credit, and pandemic), validation across additional regimes would further test generality. The realism metric correlates generated trajectories against the single realized path per crisis, rewarding similarity to history rather than to the broader space of plausible counterfactuals; we mitigate this by also reporting severity and diversity. The mixture action (\cref{eq:mixture_distribution}) uses a moment-combined Gaussian approximation that retains multimodal expressivity during training but samples unimodally per step, and richer per-step samplers are a natural extension. Finally, like any oracle-guided method, \MethodName inherits the oracle's blind spots, though our ablations show graceful degradation as oracle quality declines. \textbf{Future work} will extend \MethodName to additional domains and to neural architecture search and hyperparameter optimization.

\section*{Reproducibility}
All hyperparameters are listed in \cref{tab:hyperparams}, and all reported results are averaged over 30 independent seeds. The macro-financial dataset is constructed from public sources (FRED, Yahoo Finance) using the strict temporal train/test protocol of \cref{subsec:ExperimentalSetup}.

\section*{Use of AI Assistance}
A large language model was used for copy-editing (grammar and phrasing) only, not for generating scientific content or code.

\IEEEtriggeratref{23}
\bibliographystyle{IEEEtran}
\bibliography{example_paper}

\end{document}